\documentclass[11pt]{article}
\usepackage{amssymb}
\usepackage{amsmath}
\usepackage{algorithm}
\usepackage{algorithmic}
\usepackage{pifont}   
\newcommand{\cmark}{\ding{51}}%
\newcommand{\xmark}{\ding{55}}%
\usepackage[final]{acl}

\usepackage{times}
\usepackage{latexsym}

\usepackage[T1]{fontenc}

\usepackage[utf8]{inputenc}

\usepackage{microtype}

\usepackage{inconsolata}

\usepackage[table]{xcolor}
\usepackage{booktabs}  
\usepackage{multirow}  
\usepackage{graphicx}  
\definecolor{rowgray}{gray}{0.92} 

\definecolor{rowpurple}{rgb}{0.7216, 0.8706, 0.9569}
\title{From AR to Diffusion: Efficiently Adapting Large Language Models with Strictly Causal and Elastic Horizons}

\author{
 \textbf{Xiangyu Ma\textsuperscript{1}\thanks{These authors contributed equally to this work.}},
 \textbf{Teng Xiao\textsuperscript{2}\footnotemark[1]},
 \textbf{Zuchao Li\textsuperscript{1}\thanks{Corresponding Author. This work was supported by the National Natural Science Foundation of China (No. 62306216), the Technology Innovation Program of Hubei Province (No. 2024BAB043).}},
 \textbf{Lefei Zhang\textsuperscript{2}}
\\
\\
 \textsuperscript{1}School of Artificial Intelligence, Wuhan University, Wuhan, China,\\
 \textsuperscript{2}School of Computer Science, Wuhan University, Wuhan, China \\
 \texttt{\{maxiangyu, xiaoxiao, zcli-charlie, zhanglefei\}@whu.edu.cn}
}

\begin{document}
\maketitle
\begin{abstract}
Diffusion models promise efficient parallel text generation but rely on bidirectional attention, creating a structural mismatch with pre-trained Autoregressive (AR) models. This incompatibility precludes reusing robust AR priors, necessitating prohibitive pre-training from scratch. To bridge this gap, we propose FLUID, a framework that efficiently adapts AR backbones to the diffusion paradigm. By enforcing Strictly Causal Alignment, FLUID enables seamless initialization from standard GPT-style checkpoints, circumventing the need for massive pre-training. Furthermore, we introduce Elastic Horizons, an entropy-driven mechanism that dynamically modulates denoising strides based on local information density rather than fixed schedules. Experiments demonstrate that FLUID achieves state-of-the-art performance while reducing training costs by orders of magnitude, effectively reconciling established AR foundations with efficient parallel generation. Our code is available at \url{https://huggingface.co/MYTH-Lab/FLUID}.
\end{abstract}


\section{Introduction}
\label{sec:intro}
Autoregressive (AR) language models, trained via the next-token prediction paradigm, constitute the cornerstone of modern natural language processing. By conditioning each token solely on its preceding context, AR models ensure rigorous logical consistency and training stability, powering breakthroughs in reasoning-intensive tasks such as code generation and mathematics~\citep{lmis}. However, this strictly sequential formulation imposes a bottleneck: inference latency grows linearly with sequence length~\citep{shiluohe,tangzicong,zhaoyi}. As model scales and context windows expand, this serial decoding cost increasingly dominates deployment budgets, creating an urgent demand for parallel generation paradigms~\citep{yanghaoqi,chenjingli}.

To transcend this limitation, Discrete Diffusion Models (DLMs) have emerged as a promising alternative, offering the capability to generate multiple tokens in parallel through iterative denoising~\citep{ssd-lm}. Approaches such as LLaDA~\citep{llada} and Dream~\citep{dream} demonstrate that diffusion can model global context effectively. However, standard diffusion models typically rely on bidirectional attention mechanisms. While effective for global coherence, this architecture creates a critical structural mismatch with the pre-trained priors of ubiquitous AR models (e.g., Llama, GPT). This incompatibility precludes the efficient reuse of existing checkpoints, necessitating computationally prohibitive pre-training from scratch.

\begin{figure}[t]
  \includegraphics[width=0.9\linewidth]{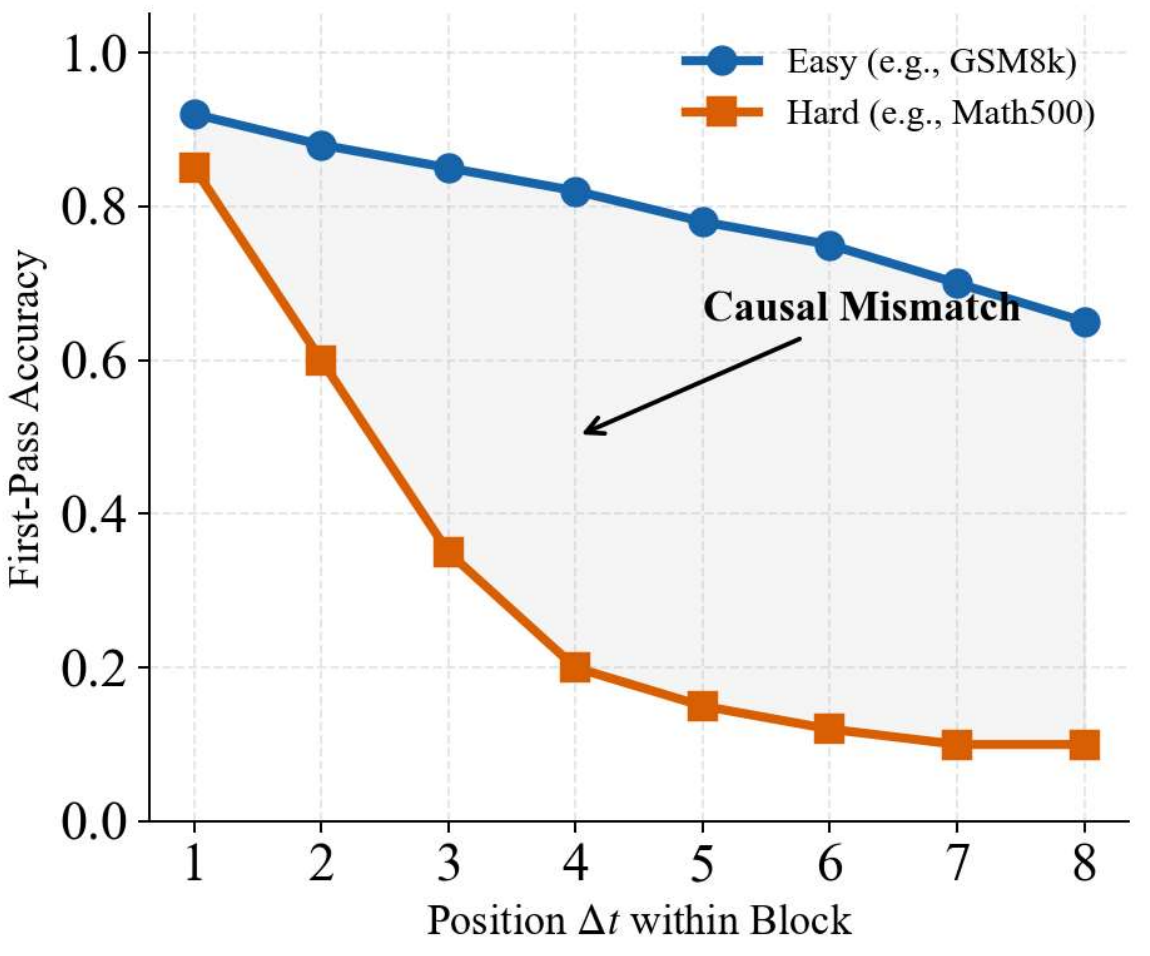}
  \centering
  \caption{Causal mismatch in fixed-size block diffusion. First-pass accuracy decays significantly faster for high-entropy data (e.g., complex reasoning steps in MATH500) than low-entropy text (e.g., GSM8k) as lookahead increases.}
  \label{fig:motivation}
\end{figure}

Recent ``Block Diffusion'' strategies~\citep{BlockDiffusion,sdar} attempt to bridge this gap by adopting a ``semi-autoregressive'' hybrid design: sequences are partitioned into fixed blocks, inheriting AR's serial dependency between blocks while applying bidirectional diffusion within them. While this mitigates some efficiency bottlenecks, we identify a critical limitation in the fixed-size nature of these strategies.

Despite these advancements, we identify a critical limitation in fixed-size block diffusion: the \textit{Entropy-Horizon Dilemma}. 
As illustrated in Figure \ref{fig:motivation}, rigid generation windows fundamentally misalign with the dynamic entropy of natural language. Text exhibits varying information density, oscillating significantly between deterministic functional phrases and high-uncertainty reasoning steps.
In high-entropy regions (e.g., MATH500), large blocks cause a rapid decay of the ``causal horizon,'' necessitating aggressive error correction that negates parallelization gains. Conversely, in low-entropy regions (e.g., GSM8K), conservative blocks fail to fully exploit parallel decoding potential. This static approach decouples generation cadence from the intrinsic semantic rhythm, limiting both inference speed and quality.

To resolve these structural and dynamic mismatches, we propose FLUID (Flexible Unidirectional Inference Diffusion), a novel framework designed to efficiently adapt AR models into parallel diffusers. Unlike traditional diffusion models that enforce bidirectional dependencies, FLUID introduces Strictly Causal Alignment by imposing a unidirectional attention mask during the diffusion process. This design realigns the generative mechanism with the inductive biases of pre-trained AR backbones, enabling seamless initialization and efficient fine-tuning without the need for massive pre-training. Furthermore, we replace fixed block boundaries with Elastic Horizon Modeling, a mechanism driven by real-time entropy estimation. This allows FLUID to dynamically modulate the generation window—accelerating through high-confidence segments while allocating dense computation to complex reasoning steps—thereby achieving a seamless balance between inference speed and generation quality.




    

Comprehensive Experiments across general understanding, mathematical reasoning, and code generation demonstrate that FLUID matches the performance of top-tier autoregressive models while significantly outperforming existing bidirectional diffusion baselines. By enforcing a unidirectional attention mask to realign with the inductive biases of pre-trained backbones, our framework validates strictly causal diffusion as a paradigm that superiorly balances training efficiency, inference latency, and logical consistency. This alignment—enabling seamless initialization from standard GPT-style checkpoints to reduce training costs by orders of magnitude—theoretically facilitates native KV cache support, establishing a decisive efficiency advantage over non-linear bidirectional diffusers. Leveraging an entropy-driven Elastic Horizon mechanism, FLUID dynamically modulates its generation window, accelerating through predictable segments while ensuring high-fidelity reasoning in complex transitions to effectively mitigate the semantic fracture inherent in fixed-size strategies. Ultimately, our work establishes a scalable foundation for transforming established AR foundations into efficient parallel diffusers without compromising structural integrity or logical coherence.

\begin{figure*}[t]
  \centering
  \includegraphics[width=6in]{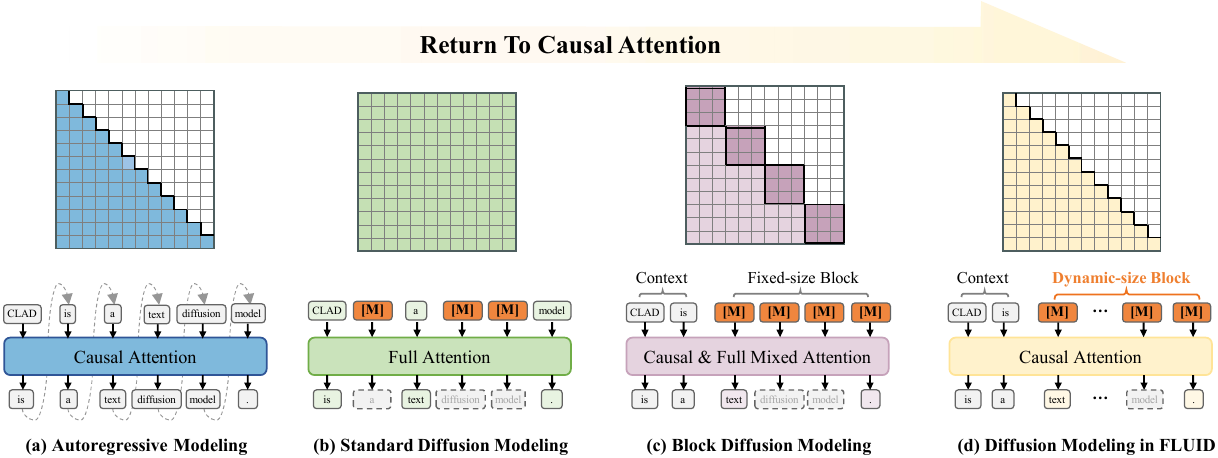}
  \caption{From Autoregressive to FLUID: Evolution of four language model generation paradigms. The diagram highlights the pivotal transition to FLUID (d), which integrates Unidirectional Attention to maintain strictly causal alignment with AR priors, and Dynamic Mask Blocks to adaptively modulate generation horizons based on sequence entropy, resolving the structural and dynamic mismatches inherent in prior paradigms.}
  \label{fig:paradiams}
\end{figure*}
\section{Related Work}

\subsection{Discrete Diffusion for Language Modeling}

DLMs offer a non-autoregressive alternative to traditional modeling by iteratively denoising discrete tokens \citep{austin2021structured, li2022diffusion}. Utilizing masking or absorbing states, DLMs enable parallel generation and mitigate exposure bias \citep{zhou2024diffusion, zeng2025treediff}. While initially computationally demanding \citep{gulrajani2023search}, recent Masked Diffusion Models have bridged this efficiency gap. \citet{lou2024discrete} matched GPT-2 performance, and subsequent scaling to the billion-parameter level—exemplified by LLaDA \citep{llada} and Mercury Coder \citep{khanna2025mercury}—has achieved parity with strong AR baselines like LLaMA3 \citep{llama3}, establishing DLMs as a scalable foundation for language generation.

\subsection{Adapting Autoregressive Models to Diffusion} 


To circumvent prohibitive pre-training costs, recent research adapts pre-trained AR backbones to diffusion, leveraging robust priors to reframe sequential generation as parallel denoising.
DiffuLLaMA \citep{diffullama} pioneered this by relaxing causal masking and applying parameter-efficient fine-tuning. Subsequent hybrid models like SDAR \citep{sdar} integrate AR consistency with diffusion refinement to ensure scalable quality. Meanwhile, efficiency-focused methods such as Fast-DLLM \citep{wu2025fast} and Fast-DLLM-v2 \citep{wu2025fastv2} mitigate inference bottlenecks via training-free acceleration. Collectively, these works validate the computational efficiency of transforming AR foundations into robust diffusion decoders.


\subsection{Block-Wise Adaptation and Hybrid Approaches} 

While capturing global context, full-sequence diffusion incurs high computational overhead and misaligns with the causal inductive biases of AR pre-training. Although recent caching mechanisms \citep{dLLMCache, dkv_cache, wu2025fast} attempt to alleviate these inefficiencies, they do not fundamentally alter the generation paradigm. 
To enable scalable generation, Semi-Autoregressive (or Block-Wise) approaches \citep{ssd_llm, BlockDiffusion} partition sequences into segments, denoising the current block given committed priors. While balancing quality and speed, these methods typically rely on rigid boundaries and bidirectional attention, restricting dynamic adaptability.

\section{Preliminaries}
\label{sec:preliminaries}

Consider a sequence of tokens $\mathbf{x} = [x_1, \dots, x_L]$ from a vocabulary $\mathcal{V}$. Let $\mathcal{D}$ denote the data distribution. We aim to learn a model $p_\theta$ that approximates $\mathcal{D}$.

\begin{figure*}[t]
  \centering
  \includegraphics[width=\textwidth]{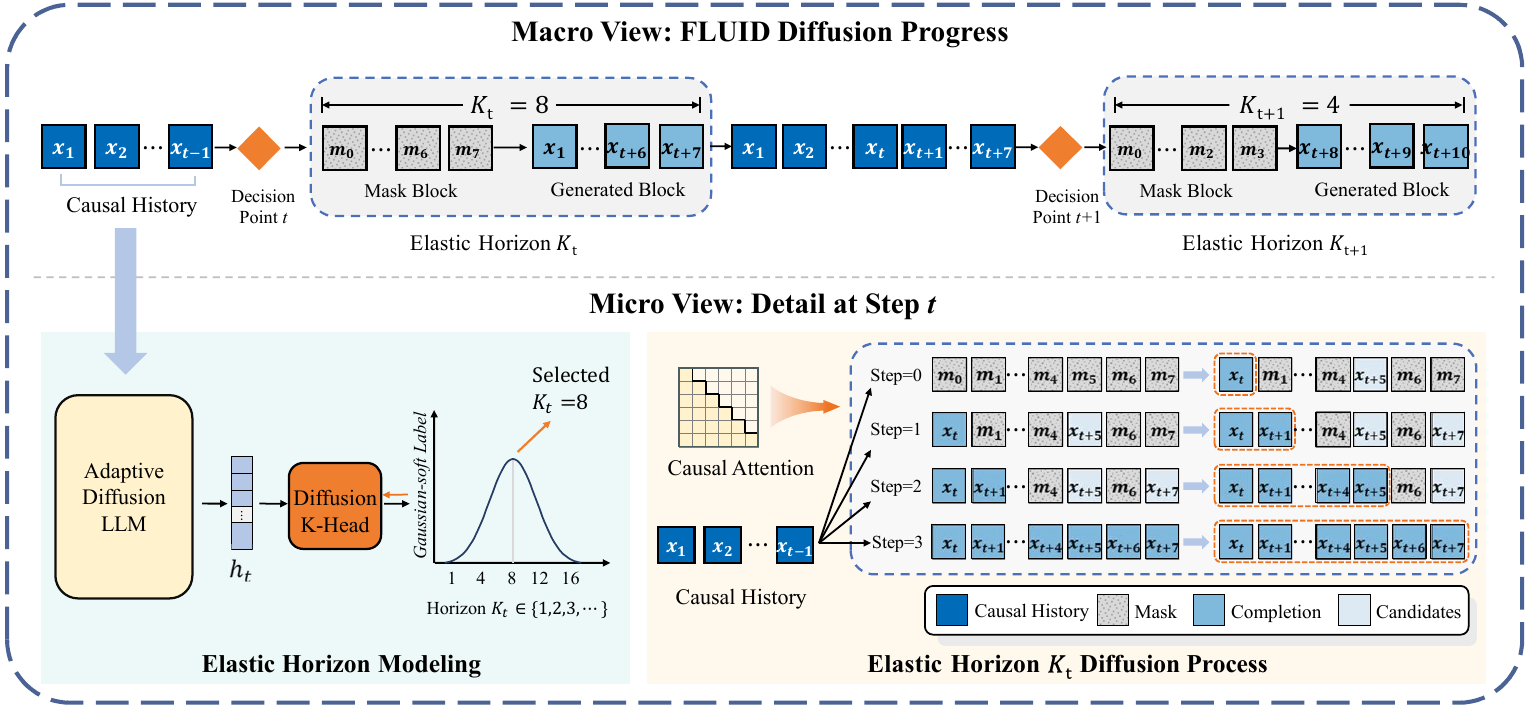}
  \caption{The Schematic Framework of FLUID. The Macro View (Top) depicts the Adaptive Diffusion Progression, where the model dynamically adjusts the lookahead horizon $K_t$. The Micro View (Bottom) details the mechanism at step $t$: the Elastic Horizon Modeling derives the optimal window size from hidden states to guide the Elastic Horizon Diffusion Process, which iteratively refines masked tokens under strict causal constraints.}
  \label{fig:overview}
\end{figure*}

\subsection{Autoregressive Modeling}
Standard AR models decompose the joint probability distribution into a product of conditional probabilities via the chain rule:
\begin{equation}
    p_{\text{AR}}(\mathbf{x}) = \prod_{i=1}^L p_\theta(x_i \mid \mathbf{x}_{<i}),
    \label{eq:ar_objective}
\end{equation}
where $\mathbf{x}_{<i}$ denotes the history tokens preceding position $i$. Crucially, this factorization enforces a \textit{strictly causal dependency} structure.

\subsection{Discrete Diffusion Modeling}
\label{ssec:discrete_diffusion}
To enable non-autoregressive generation, DLMs reframe sequence generation as a corruption-and-denoising process.

\paragraph{Forward Process.}
The forward process $q(\mathbf{x}_t \mid \mathbf{x}_0)$ progressively corrupts the clean data $\mathbf{x}_0$ over continuous time $t \in [0, 1]$. Following the absorbing state mechanism \cite{austin2021structured}, tokens are independently masked based on a noise schedule $\alpha_t$. The transition probability for a token $x_i$ is:
\begin{equation}
\begin{aligned}
    q(x_{t,i} \mid x_{0,i}) = &\ \alpha_t \mathbb{I}(x_{t,i} = x_{0,i}) \\
    &+ (1 - \alpha_t) \mathbb{I}(x_{t,i} = \texttt{[M]}),
\end{aligned}
\end{equation}
where \texttt{[M]} denotes the mask token. As $t \to 1$, the sequence converges to a fully masked state.

\paragraph{Reverse Process and Optimization.}
To reverse this corruption of generative process, following \citet{ye2025dream}, we optimize a weighted cross-entropy objective as a tractable surrogate for the variational lower bound (ELBO):

\begin{equation}
\begin{aligned}
    \mathcal{L}(\theta) = -\mathbb{E}_{t, \mathbf{x}_0, \mathbf{x}_t} \bigg[ & w(t) \sum_{i=1}^L \mathbb{1}_{[x_{t,i}=\texttt{[M]}]} \\
    & \log p_\theta(x_{0,i} \mid \mathbf{x}_t) \bigg],
\end{aligned}
\label{eq:dream_loss}
\end{equation}
where $w(t)$ is a time-dependent weighting term.

Typically, the denoiser $p_\theta$ uses bidirectional attention to model global context $\mathbf{x}_t$~\cite{ye2025dream}.

\subsection{Block-Wise Semi-Autoregressive Strategies}
\label{ssec:block_diffusion}

As shown in Figure~\ref{fig:paradiams}, Block Diffusion strategies \cite{BlockDiffusion} bridge AR and diffusion by partitioning sequences into blocks $\mathbf{x} = [\mathbf{x}^1, \dots, \mathbf{x}^B]$. This approach factorizes the likelihood autoregressively across blocks while employing diffusion:
\begin{equation}
    p_\theta(\mathbf{x}) = \prod_{b=1}^B p_\theta(\mathbf{x}^b \mid \mathbf{x}^{<b}).
\end{equation}
In conventional settings, blocks have a fixed size $L_{\text{block}}$, and intra-block generation retains bidirectional visibility.

\section{FLUID}
\label{sec:method}
\subsection{Strictly Causal Diffusion Backbone}
\label{ssec:causal_realignment}

Building upon the discrete diffusion framework (\S\ref{ssec:discrete_diffusion}; Figure~\ref{fig:overview}), FLUID enforces a \textit{strictly causal} dependency structure. Departing from standard bidirectional implementations that model global noisy context~\citep{ye2025dream,nie2024scaling}, we impose a structural constraint to preserve the autoregressive inductive bias of pre-trained LLMs.

Formally, we inject a lower-triangular attention mask $\mathbf{M}$ into the Transformer. For a query at position $i$ and key at position $j$, the attention scores are modulated as:
\begin{equation}
    \text{Attention}(i, j) = 
    \begin{cases} 
    \frac{\mathbf{q}_i \mathbf{k}_j^\top}{\sqrt{d_k}} & \text{if } j \le i, \\
    -\infty & \text{otherwise}.
    \end{cases}
    \label{eq:causal_mask}
\end{equation}
This constraint restricts the conditional probability of restoring token $x_i$ to depend solely on the history $\mathbf{x}_{t,<i}$, effectively pruning all connections to acausal future positions.

\subsection{Elastic Horizon Modeling}
\label{ssec:elastic_horizon}

Standard fixed-size blocks neglect the variable entropy of natural language, fracturing semantic units in high-uncertainty regions while wasting cycles in deterministic ones. We resolve this \textit{Entropy-Horizon Dilemma} via \textit{Elastic Horizons}, dynamically modulating the generation stride $K_t$ based on local confidence.

\subsubsection{Probabilistic Horizon Estimator}
We append a lightweight \textit{Diffusion K-Head} to the frozen backbone to predict the optimal stride. Unlike scalar regression, we model the horizon as a probability distribution to capture the inherent ambiguity of semantic boundaries. Formally, the K-Head maps the final hidden state $h_t$ to a categorical distribution over $k \in \{1, \dots, K_{\max}\}$:
\begin{equation}
    \begin{aligned}
    \mathbf{z}_t &= \text{MLP}(h_t), \\
    P_\phi(k \mid h_t) &= \text{Softmax}(\mathbf{z}_t)_k.
    \end{aligned}
\end{equation}
This probabilistic formulation enables the model to express uncertainty, favoring conservative steps when boundaries are ambiguous.

\subsubsection{Competence Boundary Supervision}
\label{sssec:competence_supervision}
In the absence of explicit labels, we frame horizon prediction as learning a \textit{Competence Boundary}. We define the oracle horizon $K^*_t$ as the maximum span of high-confidence generation, derived from the probed future loss sequence $\mathcal{L}$:
\begin{equation}
    K^*_t = \max \left\{ k \mid \frac{1}{k} \sum_{j=1}^k \mathcal{L}_{t+j} < \tau \right\}.
\label{eq:k*_t}
\end{equation}
To reflect the ordinal nature of this boundary, we supervise the K-Head using a Gaussian soft target $\mathcal{Q}$ centered at $K^*_t$:
\begin{equation}
    \mathcal{Q}(k) \propto \exp\left( -\frac{(k - K^*_t)^2}{2\sigma^2} \right).
\end{equation}
Minimizing the KL divergence $D_{\text{KL}}(\mathcal{Q} \parallel P_\phi)$ compels the estimator to align its planning horizon with the backbone's intrinsic generative capability.

\begin{figure}[t]
  \centering
  \includegraphics[width=3in]{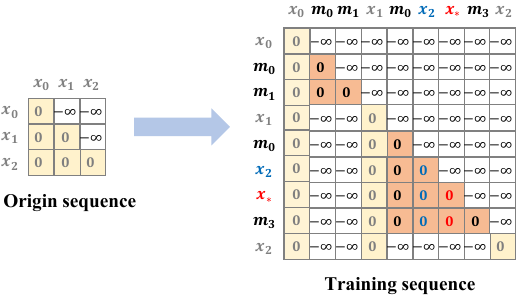}
  \caption{Illustration of Training Objective.
    The original sequence is transformed via dynamic expansion. Block $x_1$ ($k=4$) exhibits a mixed state of stochastically restored ground-truth ($x_2$) and injected noise ($x_*$), governed by a strictly causal attention matrix.}
  \label{fig:training_process}
\end{figure}

\subsection{Training}
\label{sec:training}

Training follows a two-stage curriculum: causal denoising, then horizon calibration.

\subsubsection{Stage I: Joint Causal Backbone Training}

Stage I fine-tunes the backbone $\theta$ with the K-Head frozen, employing a hybrid objective that synergizes autoregressive generation with masked denoising to preserve linguistic priors.

As depicted in Figure~\ref{fig:training_process}, for each sequence $\mathbf{x}$, we sample a horizon $K \sim \mathcal{U}[1, K_{\max}]$ to partition the input into history $\mathbf{x}_{\text{obs}}$ and targets $\mathbf{x}_{\text{mask}}$. To enhance robustness, we apply \textit{Stochastic Restoration} by injecting $10\%$ noise into the masks (\S\ref{ssec:causal_realignment}).

The optimization minimizes a hybrid loss function:
\begin{equation}
    \begin{aligned}
    \mathcal{L}_{\text{Stage1}} = & \, \underbrace{-\sum_{x_i \in \mathbf{x}_{\text{obs}}} \log p_\theta(x_i \mid \mathbf{x}_{<i})}_{\mathcal{L}_{\text{AR}}} \\
    & \hspace{-1em} + \underbrace{\mathbb{E}_{t} \left[ -\sum_{x_j \in \mathbf{x}_{\text{mask}}} w_t \log p_\theta(x_j \mid \mathbf{x}_t) \right]}_{\mathcal{L}_{\text{Diff}}}.
    \end{aligned}
\end{equation}
The $\mathcal{L}_{\text{AR}}$ term maintains the semantic stability of the prefix, while $\mathcal{L}_{\text{Diff}}$ drives the parallel denoising of the $K$-token span under strict causal constraints.

\subsubsection{Stage II: Probabilistic Horizon Training}
\label{sssec:stage2}

In Stage II, we freeze the backbone to exclusively optimize the K-Head parameters $\phi$, framing horizon estimation as a distribution matching task.

We probe the frozen backbone by inserting the maximum mask span ($K_{\max}$) to elicit its intrinsic uncertainty boundaries. Based on the competence boundary $K^*$ derived in Eq.~\ref{eq:k*_t}, we construct a Gaussian soft target $\mathcal{Q}$.

The K-Head is trained to align its predicted horizon distribution $P_\phi$ with the target $\mathcal{Q}$ by minimizing the Kullback-Leibler divergence:
\begin{equation}
    \mathcal{L}_{\text{Stage2}} = D_{\text{KL}}\left( \mathcal{Q} \parallel P_\phi(\cdot \mid h_t) \right).
\end{equation}
This objective calibrates FLUID to learn a smooth, ordinal representation of its capabilities, inducing conservative strides in high-entropy contexts.
\begin{algorithm}[t]
\small 
\caption{Dynamic Causal Diffusion Inference}
\label{alg:inference}
\begin{algorithmic}[1]
\REQUIRE Model $p_\theta$, K-Head $f_\phi$, Prompt $x$, Threshold $\gamma$, Limit $L$
\WHILE{length($x$) $< L$ and not EOS}
    \STATE $h \gets \text{Encoder}(x)$; \quad $K \gets \text{argmax} f_\phi(h)$ \COMMENT{Plan Horizon}
    \STATE $Y \gets [M]^K$; \quad $step \gets 0$
    \WHILE{$step < K$}
        \STATE $P \gets p_\theta([x; Y])$ \COMMENT{Parallel Denoising Sweep}
        \STATE $Y[step] \gets \text{argmax}(P[step])$ \COMMENT{Force Update}
        \STATE $\Delta \gets 0$; \quad $chain\_valid \gets \text{True}$
        \FOR{$j \gets step+1$ \TO $K-1$}
            \IF{$\max(P[j]) > \gamma$} 
                \STATE $Y[j] \gets \text{argmax}(P[j])$ \COMMENT{Confidence Gating}
                \IF{$chain\_valid$}
                \STATE $\Delta \gets \Delta + 1$
                \ENDIF
            \ELSE 
                \STATE $chain\_valid \gets \text{False}$
            \ENDIF
        \ENDFOR
        \STATE $x \gets [x; Y[step : step+\Delta]]$ \COMMENT{Commit Valid Chain}
        \STATE $step \gets step + 1 + \Delta$ \COMMENT{Adaptive Stride}
    \ENDWHILE
\ENDWHILE
\RETURN $x$
\end{algorithmic}
\end{algorithm}
\subsection{Dynamic Causal Inference}
\label{ssec:inference}

FLUID reframes inference as a dynamic causal diffusion process, orchestrating an interplay between entropy-aware planning and parallel denoising (Algorithm~\ref{alg:inference}). 

At step $t$, the K-Head projects the causal hidden state $h_t$ to determine the horizon $K_t = \operatorname*{argmax}_k P_\phi(k|h_t)$. This dynamically modulates the window—expanding in deterministic phases and contracting for high-entropy transitions—before initiating parallel denoising on a mask of span $K_t$.

To ensure generation quality, we apply a \textit{confidence gating} mechanism: while the immediate next token is unconditionally accepted, future predictions at relative positions $j \in \{1, \dots, K_t\}$ are accepted only if their confidence exceeds a threshold $\gamma$. The decoding cursor then advances by $1+\Delta$, where $\Delta$ denotes the length of the continuous chain of confident denoisings immediately following the current position, implicitly truncating the predicted horizon $K_t$ if an early low-confidence token interrupts the sequence.
This adaptive stride mechanism allows FLUID to ``sprint'' through coherent segments (e.g., code boilerplate) and seamlessly revert to fine-grained autoregression when entropy spikes, ensuring strictly causal correctness without the rigidity of fixed-block decoding.

\begin{table*}[t]
\centering
\renewcommand{\arraystretch}{1.1}

\resizebox{\linewidth}{!}{
\begin{tabular}{llc cc cc cc}
\toprule
& & & \multicolumn{2}{c}{\textbf{General}} & \multicolumn{2}{c}{\textbf{Math \& Science}} & \multicolumn{2}{c}{\textbf{Code}} \\
\cmidrule(lr){4-5} \cmidrule(lr){6-7} \cmidrule(lr){8-9}
\textbf{Model} & \textbf{Type} & \textbf{Tokens} & \textbf{MMLU} & \textbf{IFEVAL} & \textbf{GSM8K} & \textbf{MATH500} & \textbf{HEval} & \textbf{MBPP} \\

\midrule
LLaMA-3-8B$^\dagger$ & AR & 15T & 68.4 (5) & 49.7 (0) & 78.3 (4) & 27.4 (0) & 59.8 (0) & 57.6 (4) \\
Qwen-2.5-7B$^\dagger$ & AR & 18T & 76.6 (5) & 74.7 (0) & 91.6 (0) & 72.8 (0) & 84.8 (0) & 79.2 (4) \\
Gemma2 9B$^\dagger$ & AR & 8T & -- & -- & 76.7 (0) & -- & 68.9 (0) & 74.9 (0) \\
Deepseek 7B$^\P$ & AR & 2T & 49.4 (0) & -- & 63.0 (0) & -- & 48.2 (-) & 35.2 (-) \\

LLaDA-8B & Diff & 2.0T & 65.5 (5) & 59.9 (0) & 78.6 (4) & 36.2 (0) & 47.6 (0) & 34.2 (4) \\
LLaDA-1.5-8B & Diff & 2.0T & 66.0 (5) & 58.2 (0) & 83.3 (4) & 42.6 (0) & 52.4 (0) & 42.8 (4) \\
Dream-7B & Diff & 0.6T & 67.0 (5) & \textbf{62.5 (0)} & 81.0 (4) & 39.2 (4) & 55.5 (0) & \textbf{58.8 (4)} \\

\rowcolor{rowpurple}
\textbf{FLUID-7B (Ours)} & \textbf{Diff} & \textbf{2.7B} &\textbf{67.8 (5)} & 57.7 (0) & \textbf{91.9 (4)} & \textbf{61.8 (4)} & \textbf{60.4 (0)} & 53.6 (4) \\
\bottomrule
\end{tabular}
}
\caption{Main Comparison on Standard Benchmarks. We compare FLUID against state-of-the-art Diffusion and AR baselines under Instruct settings. ``Tokens'' denotes the amount of data used for pre-training (for AR) or adaptation (for Diffusion). Note that FLUID achieves comparable performance to strong baselines using \textbf{orders of magnitude less training data} (2.7B vs. Trillions). Results indicated by $\dagger$ and \P\ are sourced from \citet{Qwen2_Audio}, \citet{qwen2_5}, and \citet{DeepSeek}. Best results in diffusion methods are \textbf{bolded}, and our results are highlighted in \colorbox{rowpurple}{\phantom{aaa}}.}
\label{tab:main_results}
\vspace{-2mm}
\end{table*}

\section{Experiments}
\subsection{Experimental Setup}
\label{sec:exp_setup}

We initialize FLUID with the openPangu-Embedded-7B~\citep{pangu} checkpoint to ensure fair comparison with baselines of equivalent scale, while capitalizing on its robust linguistic priors. 

Adaptation follows a two-stage curriculum: Stage I fine-tunes the backbone for 32,000 iterations (detail in Appendix~\ref{sec:appendix_loss}); Stage II freezes the backbone to exclusively train the Diffusion K-Head for 2,000 steps, targeting the oracle horizon $K_{t}^{*}$. For calibration, we employ a threshold $\tau = 2.8$ (Eq.~\ref{eq:k*_t}); a comprehensive sensitivity analysis is provided in the Ablation section.

\paragraph{Training Data.}
Our adaptation corpus is constructed by distilling responses from openPangu-Embedded-7B over several public instruction-tuning datasets, including \textbf{Infinity-Instruct-7M}~\citep{Infinity_Instruct_7M}, \textbf{deepctrl-sft-data} (zh)~\citep{deepctrl}, \textbf{moss-003-sft} (no-tools)~\citep{moss}, and \textbf{UltraChat}~\citep{ultraChat}. This distilled mixture covers diverse task distributions, including large-scale instruction-following data, Chinese multi-turn conversational data, open-domain dialogue, and broad high-quality instructional conversations, thereby providing broad supervision for adapting the autoregressive backbone to our strictly causal diffusion objective.

\textbf{Implementation Details.} Input sequences are set to a length of 1024, with a global batch size of 80. For parameter efficiency, we apply Rank-16 LoRA to the backbone, while the K-Head (a two-layer MLP) is the only newly introduced dense module trained from scratch. Following~\citep{BlockDiffusion}, we set $K_{max} = 16$ for our dynamic analysis. Optimization utilizes the AdamW optimizer with a learning rate of $2 \times 10^{-4}$, a 0.1 warmup ratio, and a cosine decay schedule.

\paragraph{Benchmarks.} We conduct comprehensive evaluations across three core domains. \textbf{General Knowledge and Instruction Following:} We employ MMLU~\citep{mmlu} for broad semantic understanding and IFEVAL~\citep{ifeval} to assess the model's ability to adhere to objective constraints and formatting prompts. 
\textbf{Mathematical Reasoning:} Performance is measured on GSM8K~\citep{gsm8k} for multi-step arithmetic and the more rigorous MATH500~\citep{math500} for competition-level problem solving. 
\textbf{Code Generation:} We utilize HumanEval~\citep{humaneval} and MBPP~\citep{mbpp} to evaluate the functional correctness and structural integrity of synthesized code.

\subsection{Main Results}

\noindent \textbf{Results and Analysis.} 
Table \ref{tab:main_results} compares FLUID-7B with leading AR and diffusion baselines. In reasoning, FLUID-7B achieves significant advantages, scoring 91.9 on GSM8K and 61.8 on MATH500, surpassing diffusion models like Dream-7B and LLaDA-8B by 10.9 and 13.3 points on GSM8K, respectively, and matching top AR models like Qwen-2.5-7B (91.6). These results validate our strictly causal diffusion framework. Unlike bidirectional models, which disrupt deductive chains by conditioning on noisy future contexts, FLUID’s causal masking preserves logical consistency.

FLUID-7B further excels in code generation (60.4 on HumanEval) and instruction following (57.7 on IFEval), outperforming LLaMA-3-8B-Instruct and substantially leading LLaDA-8B. This superiority is driven by the Elastic Horizon Modeling. Whereas fixed-block strategies often cause ``semantic fracture'' by truncating syntactic units, our entropy-driven mechanism dynamically modulates the generation window. By expanding for predictable syntax and contracting for high-entropy logic, FLUID ensures structural integrity while bridging the gap between diffusion-based planning and AR-style stability. 

\begin{figure}[h]
  \centering
  \includegraphics[width=1.0\linewidth]{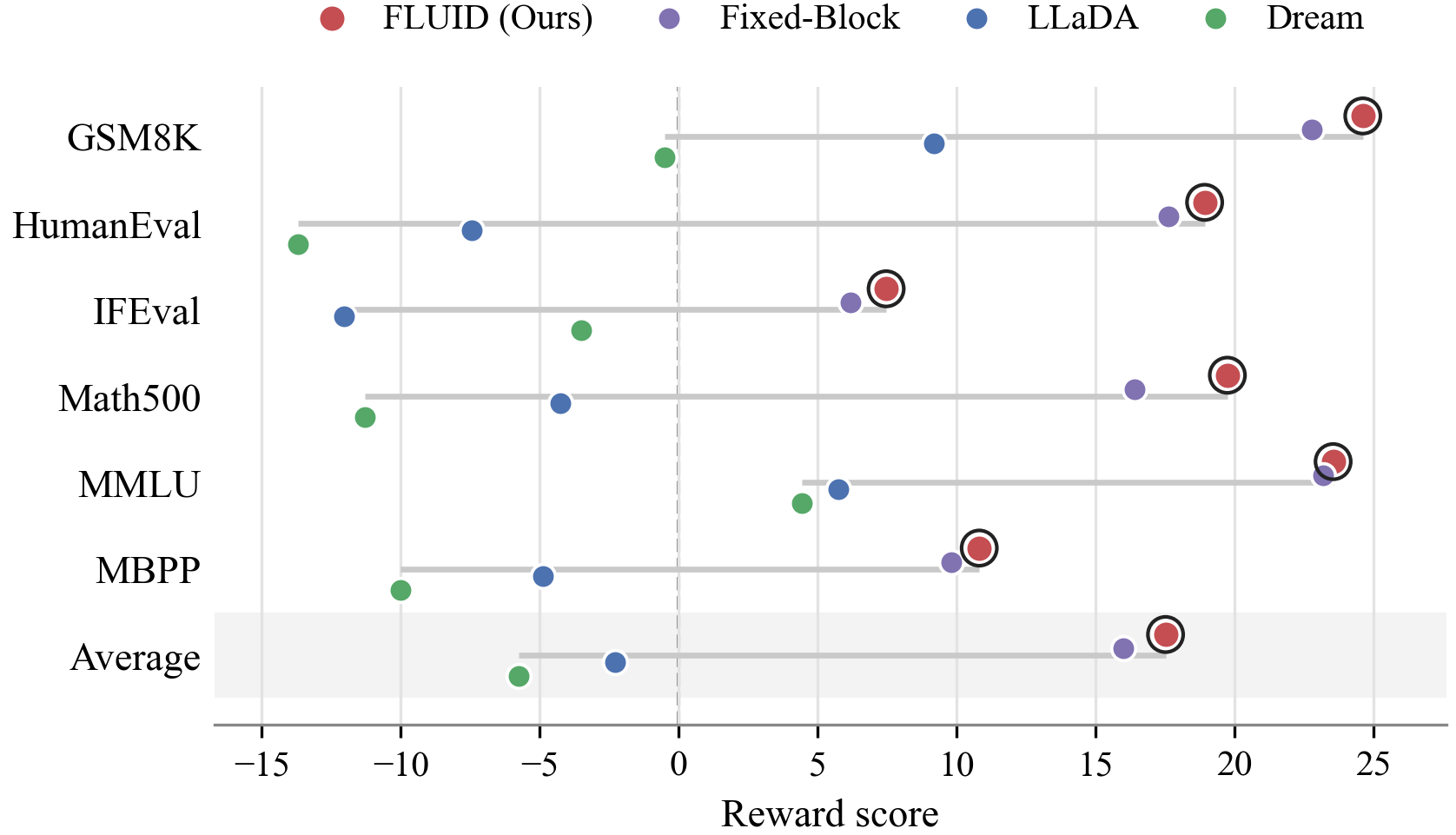}
  \caption{Semantic Quality Evaluation via Skywork-Reward-V2.}
    \label{fig:radar_evaluation}
\end{figure}

\subsection{Semantic Quality Evaluation}

Beyond surface-level metrics, we assess generation quality using Skywork-Reward-V2~\citep{skywork}, a preference model designed to evaluate intrinsic helpfulness and logical coherence. As shown in Figure~\ref{fig:radar_evaluation}, FLUID achieves the highest reward scores across all five domains, validating its superior alignment with human intent.

Notably, FLUID significantly outperforms bidirectional diffusion baselines (e.g., LLaDA, Dream), particularly in reasoning-intensive tasks like GSM8K and MATH500. This confirms that strictly causal attention preserves deductive chains often disrupted by the noisy, acausal contexts inherent to standard diffusion models. Furthermore, FLUID maintains a consistent lead over Fixed-Block strategies. By replacing rigid boundaries with entropy-driven Elastic Horizons, our model effectively mitigates ``semantic fracture''—ensuring structural integrity in code generation and continuity in complex reasoning without incurring the waiting costs of conservative blocking.

\begin{figure*}[t]
  \centering
  \includegraphics[width=1.0\linewidth]{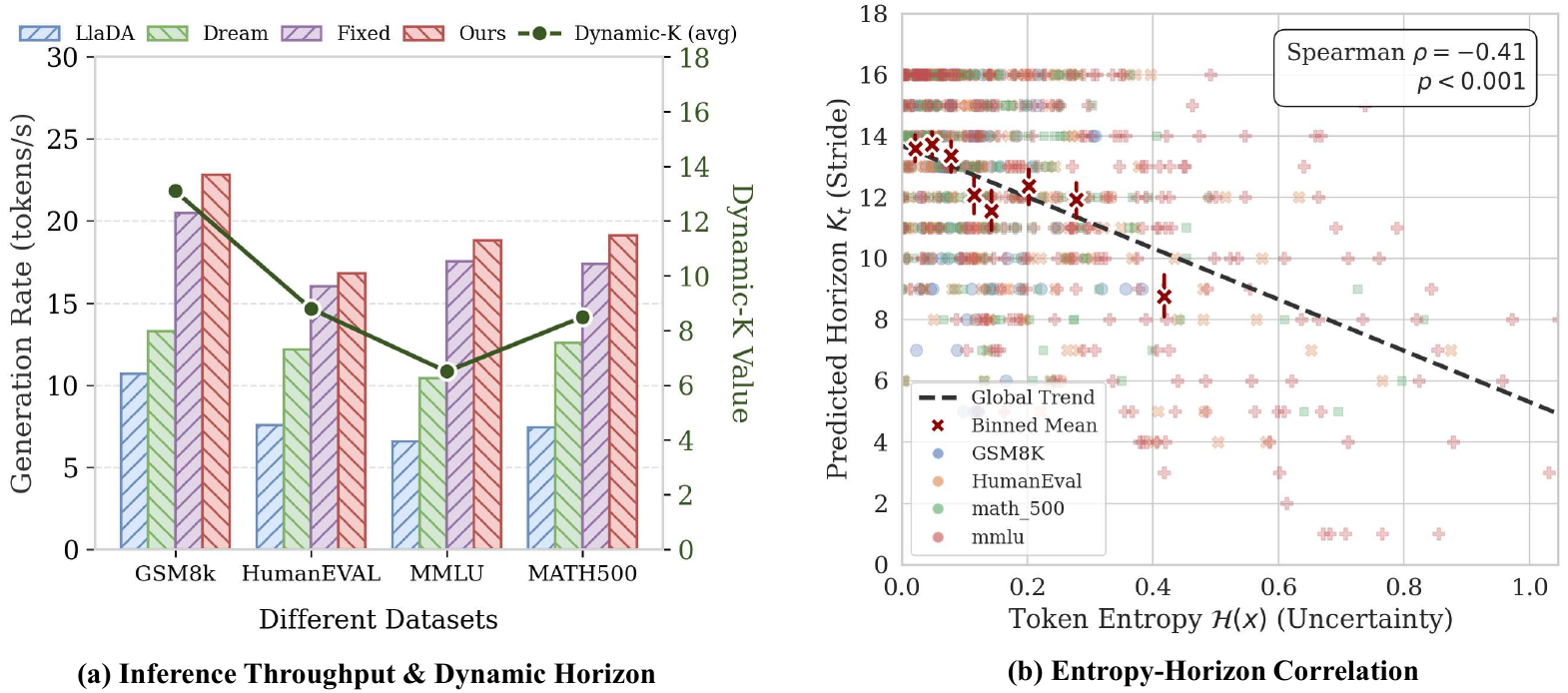}
  \caption{Efficiency and Accuracy Comparison.
    (a) All models are evaluated without KV cache or FlashAttention to ensure a fair architectural comparison. (b) Verification of K-Head Accuracy.}
  \label{fig:efficiency_analysis}
\end{figure*}

\subsection{Ablation Study}
In this section, we analyze the effectiveness of the proposed adaptation methodologies.

\textbf{Impact of Causal Attention}. We evaluate the necessity of strictly causal masking. As shown in Table \ref{tab:ablation}, the bidirectional attention baseline significantly underperforms, lagging behind FLUID by 9.4\% on GSM8K. This confirms that conditioning on noisy, acausal future contexts disrupts the pre-trained autoregressive logic chain, leading to incoherent reasoning. By enforcing strict causal masking, FLUID realigns the diffusion process with the backbone's AR priors, effectively restoring the reasoning capabilities essential for complex tasks.

\textbf{Benefit of Elastic Horizon}. We further compare FLUID against the Fixed Block Size ($k=16$) baseline. While fixed blocks improve stability over standard diffusion, they still suffer from \textit{semantic fracture}, the arbitrary truncation of coherent logic units—resulting in a 5.5\% deficit on HumanEval. FLUID's adaptive horizon addresses this by dynamically expanding generation windows based on local entropy. This ensures that atomic syntactic units (e.g., function definitions) remain intact, yielding superior performance in structure-sensitive tasks like code generation.


\begin{table}[h]
\centering
\small
\setlength{\tabcolsep}{1pt}
\begin{tabular}{lccccc}
\toprule
& \multicolumn{2}{c}{\textbf{Components}} & \multicolumn{3}{c}{\textbf{Benchmarks}} \\
\cmidrule(lr){2-3} \cmidrule(lr){4-6}
\textbf{Method} & \textbf{Causal} & \textbf{Elastic} & \textbf{GSM8K} & \textbf{MATH500} & \textbf{HEval}\\
\midrule
Baseline & \xmark & \xmark & 82.0 & 51.2 & 42.2 \\
Baseline & \xmark & \cmark & 82.5 & 53.6 & 42.8 \\
Baseline & \cmark & \xmark & 90.6 & 59.2 & 54.9 \\
\rowcolor{rowpurple}
\textbf{FLUID (Ours)} & \cmark & \cmark & \textbf{91.9} & \textbf{61.8} & \textbf{60.4} \\
\bottomrule
\end{tabular}
\caption{Ablation study on adaptation strategies. Impact of strictly causal masking (\textit{Causal}) and entropy-aware dynamic horizons (\textit{Elastic}) relative to a bidirectional fixed-block baseline.}
\label{tab:ablation}
\end{table}

\textbf{Inference Latency.}
As shown in Figure~\ref{fig:efficiency_analysis}(a), FLUID establishes a distinct efficiency advantage. Compared to standard diffusion baselines (e.g., LLaDA, Dream) that require intensive iterative refinement, FLUID achieves approximately 2$\times$ speedup.

Crucially, FLUID outperforms the aggressive Fixed-Block ($K=16$) baseline even with a smaller average stride, highlighting the efficiency of our Elastic Horizon. On challenging tasks like MMLU, forcing large strides incurs significant ``waiting costs'' due to high entropy. Conversely, FLUID autonomously contracts its horizon (avg $K=6.5$) to prioritize robust updates. By mitigating error correction overhead, FLUID achieves higher throughput (18.82 vs. 17.52 tokens/s), demonstrating that optimal latency stems from \textit{semantic synchronization} rather than rigid block maximization.

\begin{table}[h]
\centering
\small
\setlength{\tabcolsep}{9pt}
\begin{tabular}{lccc}
\toprule
\textbf{Noise Ratio} & \textbf{GSM8K} & \textbf{MATH500} & \textbf{HEval} \\
\midrule
0\%  & 91.0 & 60.8 & 59.8 \\
5\%  & 91.3 & 61.1 & 59.8 \\
\rowcolor{rowpurple}
\textbf{10\% (Ours)} & \textbf{91.9} & \textbf{61.8} & \textbf{60.4} \\
15\% & 91.1 & 61.5 & 60.0 \\
\bottomrule
\end{tabular}
\caption{Ablation on the stochastic restoration noise ratio in Stage I training. Impact of varying proportions of injected noise within the masked span.}
\label{tab:noise_ratio_ablation}
\end{table}
\textbf{Impact of Stochastic Restoration Ratio.}
In Stage I, FLUID adopts stochastic restoration by injecting a small amount of noise into the masked span to improve robustness during causal denoising. To study its effect, we vary the noise ratio from 0\% to 15\% while keeping all other settings fixed. As shown in Table~\ref{tab:noise_ratio_ablation}, a moderate ratio gives the best trade-off, with 10\% achieving the strongest overall performance on GSM8K, MATH500, and HEval.

Without stochastic restoration (0\%), the model is trained on easier denoising patterns and is less robust to imperfect intermediate states, leading to lower performance across all benchmarks. Introducing mild noise (5\%--10\%) improves stability and better prepares the model for noisy decoding trajectories. In contrast, excessive noise (15\%) harms performance by distorting the target signal and making optimization under strict causal constraints more difficult, especially on MATH500 and HEval. Overall, these results show that a moderate restoration ratio is most effective, balancing robustness and target fidelity.

\begin{table}[h]
\centering
\setlength{\tabcolsep}{9pt}
\small
\begin{tabular}{lcccc}
\toprule
\textbf{Threshold $\tau$}  & \textbf{GSM8K} & \textbf{MATH500} & \textbf{HEval}\\
\midrule
2.6  & 90.9 & 60.4 & 59.8 \\
2.7 & 91.1 & 61.5 & 59.9 \\
\rowcolor{rowpurple} 
\textbf{2.8 (Ours)}  & \textbf{91.9} & \textbf{61.8} & \textbf{60.4} \\
 2.9 & 90.5 & 61.5 & 59.5 \\
 3.0 & 90.4 & 60.2 & 59.3 \\
\bottomrule
\end{tabular}
\caption{Ablation study on the competence boundary $\tau$.}
\label{tab:tau_sensitivity}
\end{table}

\textbf{Impact of the Competence Boundary $\tau$.}
The competence boundary $\tau$ governs the trade-off between decoding aggressiveness and causal reliability by controlling how far FLUID can safely expand its dynamic horizon. As shown in Table~\ref{tab:tau_sensitivity}, $\tau=2.8$ achieves the best overall results across GSM8K, MATH500, and HEval. When $\tau$ is set to smaller values (e.g., 2.6 or 2.7), FLUID tends to adopt a more conservative horizon. While this improves decoding stability, it also limits the benefit of parallel generation and leads to slightly weaker overall performance. In contrast, larger values (e.g., 2.9 or 3.0) encourage more aggressive horizon expansion, but also increase error correction costs once the model overestimates its confidence, especially on reasoning-intensive tasks such as MATH500.

Overall, the results exhibit a clear bell-shaped trend centered at $\tau=2.8$. This suggests that an appropriate competence boundary is crucial for balancing decoding efficiency and causal consistency, and confirms that $\tau=2.8$ provides the most effective operating point for FLUID.

\subsection{Verification of K-Head Accuracy}
To verify that the Elastic Horizon reflects the model's internal confidence, we analyze the correlation between predicted horizon $K_t$ and information density. Figure~\ref{fig:efficiency_analysis}(b) reveals a significant negative correlation between uncertainty and stride length.

Task-specific strides in Figure~\ref{fig:efficiency_analysis}(a) corroborate this adaptivity. On GSM8K, where the model demonstrates high competence (91.9\% accuracy), the K-Head confidently expands the stride (average $K=13.1$) to maximize speed. Conversely, on MMLU, where decision ambiguity is higher, the horizon naturally contracts (average $K = 6.5$). This confirms that the K-Head functions as a \textit{semantic gear stick}—sprinting through high-confidence reasoning chains while downshifting to a cautious, fine-grained pace during challenging transitions.


\section{Conclusion}

In this work, we introduced FLUID, a framework that reimagines text diffusion by breaking free from the rigidity of fixed-step generation. By enforcing a strictly causal attention mechanism, FLUID bridges the structural gap between diffusion paradigms and pre-trained autoregressive priors, enabling efficient adaptation without the prohibitive costs of pre-training from scratch. Our experiments demonstrate that FLUID acts much like its namesake—dynamically modulating its generation horizon via Elastic Horizon Modeling. It flows rapidly through predictable sequences and contracts cautiously during high-entropy transitions, thereby resolving the trade-off between inference latency and reasoning integrity. Ultimately, FLUID establishes a new paradigm for efficient generative modeling, proving that aligning causal consistency with dynamic flexibility is the key to unlocking the full potential of non-autoregressive text synthesis.

\section*{Limitations}

FLUID is designed to efficiently adapt pre-trained autoregressive models into a diffusion framework. Consequently, its performance is inherently bounded by the capabilities of the source model. If the base autoregressive model suffers from hallucinations or reasoning deficits, FLUID is likely to inherit these behaviors. Furthermore, our current experiments focus primarily on adapting general-purpose LLMs (e.g., OpenPangu). The efficacy of FLUID on highly specialized domains (e.g., biomedical or legal texts) or different architectures (e.g., MoE models) has yet to be extensively verified, warranting further investigation.


\bibliography{custom}



\appendix
\section{Empirical Analysis of Decoding Trajectories in Bidirectional Diffusion}
\label{sec:appendix_trajectory}

\begin{figure*}[h]
  \centering
  \includegraphics[width=1.0\linewidth]{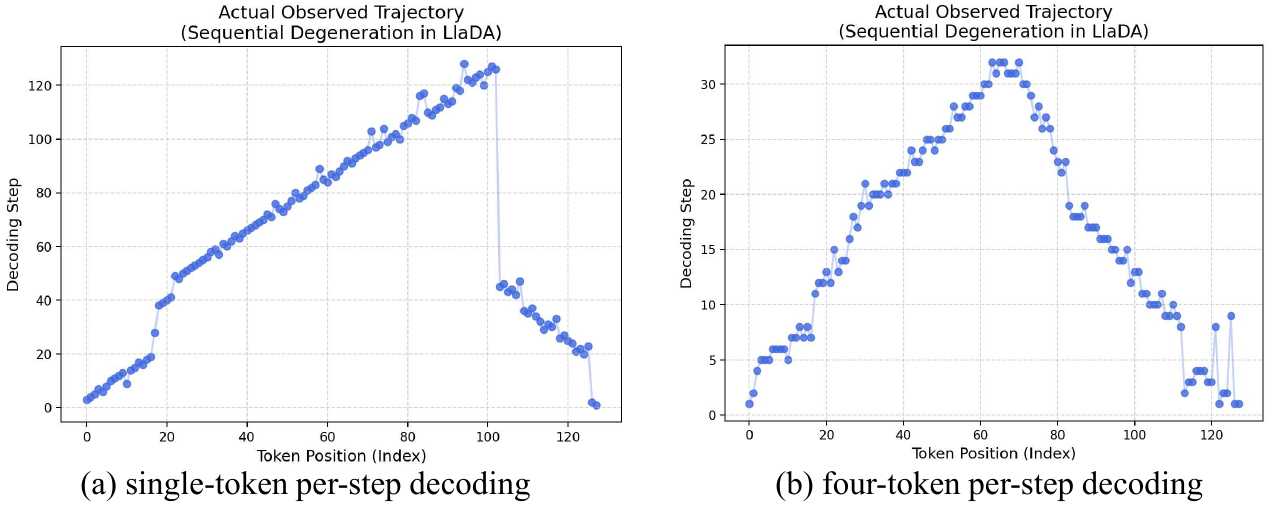}
  \caption{Decoding Trajectories of Bidirectional Diffusion. 
    (a) Incremental decoding ($K=1$) triggers an unintended sequential path, revealing a structural redundancy in bidirectional attention. 
    (b) Block-wise decoding ($K=4$) exhibits bi-terminal convergence, where non-linear filling from sequence boundaries leads to causal mismatch. 
    These behaviors substantiate FLUID's design of strictly causal alignment and adaptive horizon modulation.}
  \label{fig:diag_traj}
\end{figure*}
To further substantiate the structural and dynamic mismatches discussed in our Introduction (\S\ref{sec:intro}), we analyze the decoding trajectories of LLaDA-8B, a representative bidirectional masked diffusion model. While bidirectional attention offers theoretical global flexibility, our empirical findings reveal a "behavioral degeneration" during inference. The model’s pursuit of high-confidence sampling paths unintentionally mimics the causal constraints of AR models, without achieving their computational efficiencies.

\subsection{Sequential Degeneration in Single-Step Decoding}
When we constrain the decoding process to a single-token-per-step regime-effectively a diffusion-based greedy search---the model exhibits a nearly linear, left-to-right progression, as shown in Figure~\ref{fig:diag_traj}(a). This \textit{sequential degeneration} highlights a critical paradox: in natural language, the conditional probability $P(x_i | x_{<i})$ is typically far more deterministic than acausal lookahead predictions. Consequently, a confidence-driven remasking strategy naturally gravitates toward a causal path, rendering the bidirectional attention over future mask positions an expensive redundancy. This observation provides the empirical bedrock for our Strictly Causal Alignment in FLUID, as it aligns the model's architectural capacity with its actualized generative behavior, thereby enabling the seamless integration of KV Cache optimizations that are otherwise unattainable in bidirectional frameworks.

\subsection{Non-linear Convergence and the Causal Mismatch}
As the generation stride increases, the trajectory shifts toward a bi-terminal convergence pattern, as illustrated in Figure~\ref{fig:diag_traj}(b). The model tends to commit to both the sequence prefix and suffix tokens early in the denoising process, subsequently "filling in" the intermediate tokens. While this "pyramid" path appears to leverage global context, it introduces what we define as the \textit{Causal Mismatch}. 

Specifically, pre-sampling future tokens (e.g., sequence endings) without a coherent chain of intermediate reasoning risks \textit{semantic fracture} (\S\ref{ssec:inference})---a state where the generated "ends" are logically irreconcilable with the high-entropy reasoning steps that follow. Moreover, such non-contiguous filling precludes any spatial locality in memory access, fundamentally breaking standard caching mechanisms. This observed "filling-from-ends" behavior validates the necessity of our Elastic Horizon Modeling, which replaces this rigid, potentially incoherent global lookahead with a confidence-driven, dynamically modulated window that ensures each parallel step is supported by its causal history.

\subsection{Synthesis: From Redundancy to Fluidity}

Synthesizing these observations, it becomes evident that bidirectional diffusion models for text generation are often caught in an "architectural limbo." They carry the $O(N^2)$ computational burden of acausal attention, but either degenerate into sequential paths for quality or adopt non-linear paths that jeopardize logical consistency. By reframing the diffusion process within a strictly causal framework and modulating the decoding horizon elastically, FLUID resolves these contradictions, transforming the latent causal bias of LLMs from a hidden degeneration into an explicit, exploitable advantage for efficient parallel generation.

\begin{figure}[htbp]
    \centering
    \includegraphics[width=1.0\linewidth]{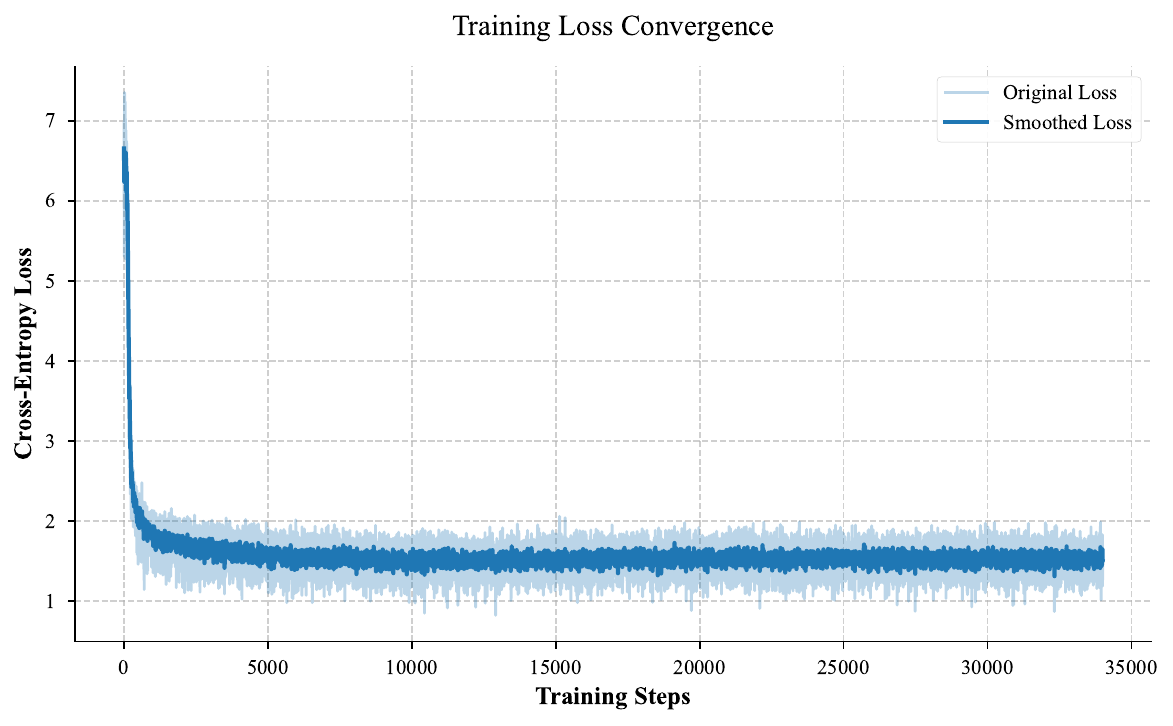}
    \caption{Training loss convergence of FLUID during Stage I. The rapid stabilization post-initialization highlights the efficiency of adapting pre-trained AR models to the diffusion paradigm via strictly causal alignment.}
    \label{fig:loss_convergence}
\end{figure}

\section{Training Dynamics and Convergence}
\label{sec:appendix_loss}

We monitor FLUID's training stability to assess the efficiency of our adaptation curriculum. Figure~\ref{fig:loss_convergence} shows the training loss (weighted cross-entropy) during the first stage of joint causal backbone training (\S\ref{sec:training}).

The plot reveals a rapid initial descent within the first 1,000 iterations, indicating effective realignment of the pre-trained openPangu-Embedded-7B priors with the causal diffusion objective. After around 10,000 steps, the loss stabilizes, reflecting mastery of the unidirectional denoising task under triangular attention constraints.

The convergence remains stable throughout the 32,000 iterations of Stage I, with the smoothed loss maintaining a plateau. This confirms that our hybrid loss function (\S\ref{sec:training}) preserves linguistic competencies while adapting to parallel generation. The subsequent 2,000 steps of Stage II, focused on calibrating the Diffusion K-Head, converge quickly due to the frozen backbone’s established confidence boundaries.

\begin{figure*}[h]
  \centering
  \includegraphics[width=1.0\linewidth]{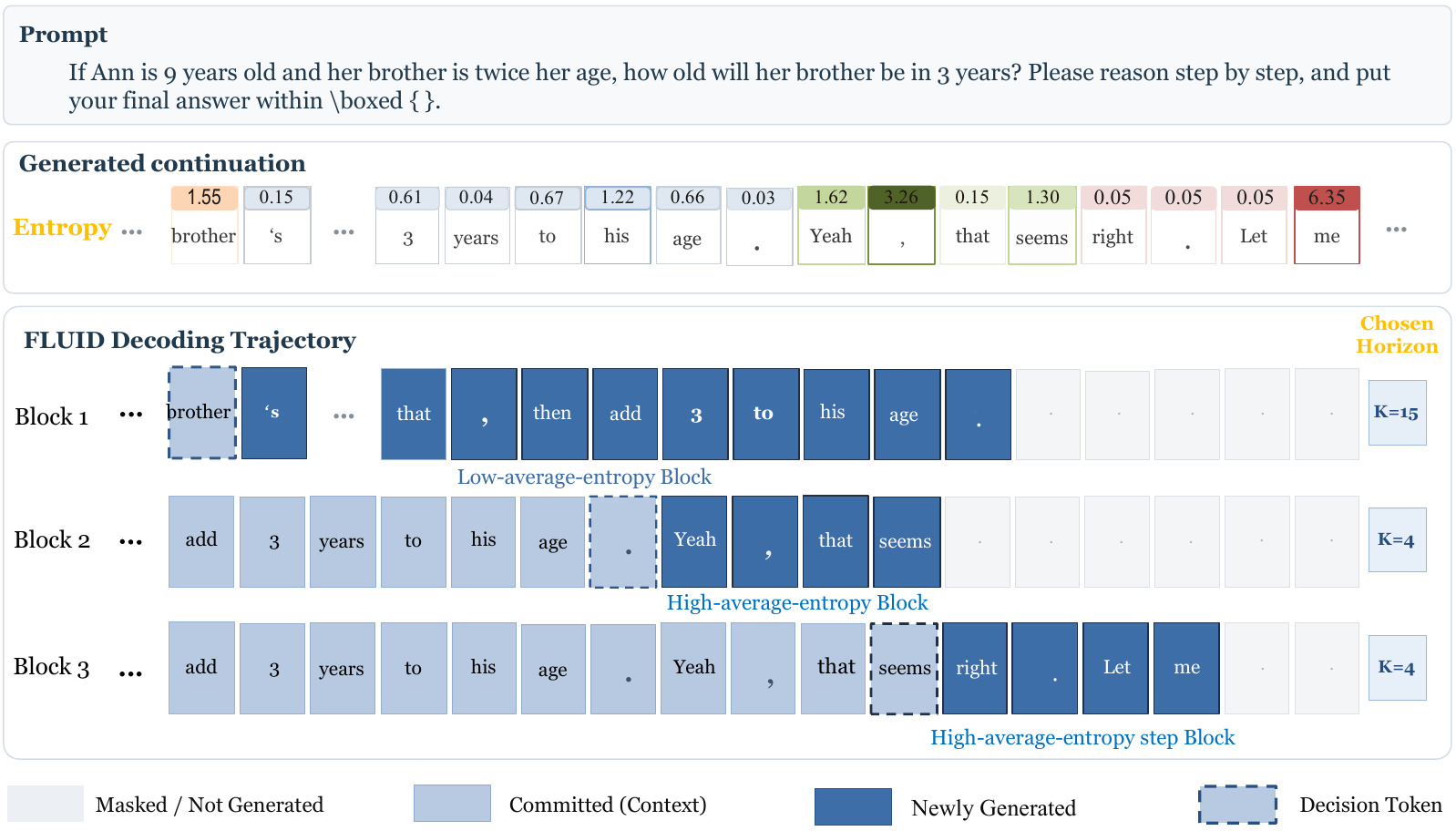}
  \caption{Visualization of FLUID's decoding trajectory, token entropy, and adaptive horizon on an arithmetic reasoning prompt.}
  \label{fig:case_study}
\end{figure*}
\section{Case Study}
To better understand how FLUID adapts its decoding horizon during reasoning, we visualize its decoding trajectory on an arithmetic example in Figure~\ref{fig:case_study}. The figure shows that FLUID does not decode with a fixed generation block. Instead, its horizon changes dynamically with local uncertainty, as reflected by the token-level entropy within the current window.

A clear low-entropy expansion and high-entropy contraction pattern can be observed. In Block~1, FLUID selects a large horizon ($K=15$) and commits a long contiguous span, since the corresponding tokens lie in a relatively stable low-entropy region and follow a predictable local reasoning template. As decoding approaches a semantic transition region, however, the entropy rises sharply around tokens such as ''Yeah'' and ''seems'', indicating increased uncertainty. FLUID therefore shrinks its horizon to $K=4$ in the following steps and switches to more conservative updates. This behavior shows that FLUID accelerates only in confident regions, while automatically slowing down near uncertain decision points.

This example highlights the key advantage of Elastic Horizons under strictly causal decoding. Natural language reasoning is heterogeneous: some spans are highly predictable, whereas others correspond to semantic transitions that require more cautious generation. Fixed-block decoding cannot adapt to such variation, and may therefore either over-commit across difficult regions or sacrifice efficiency in easy ones. In contrast, FLUID dynamically adjusts its horizon according to local uncertainty, accelerating through stable segments while preserving causal reliability near high-entropy regions. As a result, it achieves a better balance between decoding efficiency and generation quality.

\begin{table}[h]
\centering
\small
\setlength{\tabcolsep}{1pt}
\begin{tabular}{lccc}
\toprule
\textbf{Stage} & \textbf{Trained Module} & \textbf{Hardware} & \textbf{Est. Time} \\
\midrule
Stage I & LoRA Adapter & 4$\times$64GB GPUs & $\sim$80 hours \\
Stage II & K-Head & 4$\times$64GB GPUs & negligible \\
\midrule
\textbf{Total} & -- & 4$\times$64GB GPUs & $\sim$320 GPU-hours \\
\bottomrule
\end{tabular}
\caption{Computational overhead of FLUID adaptation. The total cost is approximately 320 GPU-hours, with Stage I accounting for nearly all of the training time.}
\label{tab:compute_cost}
\end{table}

\section{Training Cost Analysis}
A practical advantage of FLUID is its low adaptation cost compared with training diffusion language models from scratch. Since FLUID reuses a pretrained autoregressive backbone and only introduces lightweight adaptation modules, the overall training overhead remains modest. In our setting, the full adaptation requires approximately 320 GPU-hours in total, corresponding to about 80 wall-clock hours on 4$\times$64 GB GPUs.

As summarized in Table~\ref{tab:compute_cost}, Stage I accounts for nearly all of the training cost. In this stage, we adapt the backbone using Rank-16 LoRA for 32K iterations over approximately 2.62B tokens. By contrast, Stage II trains only a lightweight two-layer K-Head while keeping the backbone frozen, thereby introducing negligible additional overhead.

Importantly, FLUID follows the standard recipe of parameter-efficient fine-tuning rather than expensive full-model pretraining. Most backbone parameters remain frozen, and the training footprint is comparable to typical PEFT-based autoregressive fine-tuning. Therefore, FLUID can be readily scaled with standard multi-GPU training frameworks, without requiring the heavy compute budget typically associated with training diffusion language models from scratch.

\end{document}